\documentclass{article}

\usepackage{times}
\usepackage{color}
\usepackage{epsfig}
\usepackage{graphicx}
\usepackage{amsmath}
\usepackage{amssymb}
\usepackage{algorithm}
\usepackage{subfigure}

\newtheorem{thm}{Theorem}
\newtheorem{remark}{Remark}

\usepackage[pagebackref=true,breaklinks=true,letterpaper=true,colorlinks,bookmarks=false]{hyperref}
\usepackage{hyperref}
\usepackage{wasysym}

\usepackage[accepted]{icml2016}

\def\diag{\mathrm{diag}}
\def\sign{\mathrm{sign}}

\def\R{\mathbb{R}}
\def\E{\mathbb{E}}
\icmltitlerunning{FDR control and Statistical Quality Assessment of Annotators in Crowdsourced Ranking}

\begin{document}

\twocolumn[
\icmltitle{False Discovery Rate Control and Statistical Quality Assessment of Annotators in Crowdsourced Ranking}

\icmlauthor{Qianqian Xu}{xuqianqian@iie.ac.cn}
\icmladdress{State Key Laboratory of Information
Security (SKLOIS), Institute of Information Engineering, Chinese Academy of
Sciences, Beijing 100093 \& BICMR, Peking University, Beijing 100871, China}
\icmlauthor{Jiechao Xiong}{xiongjiechao@pku.edu.cn}
\icmladdress{BICMR-LMAM-LMEQF-LMP, School
of Mathematical Sciences, Peking University, Beijing 100871, China}
\icmlauthor{Xiaochun Cao}{caoxiaochun@iie.ac.cn}
\icmladdress{State Key Laboratory of Information
Security (SKLOIS), Institute of Information Engineering, Chinese Academy of
Sciences, Beijing 100093, China}
\icmlauthor{Yuan Yao}{yuany@math.pku.edu.cn}
\icmladdress{BICMR-LMAM-LMEQF-LMP, School
of Mathematical Sciences, Peking University, Beijing 100871, China}


\vskip 0.3in
]
\begin{abstract}

With the rapid growth of crowdsourcing platforms it has become easy and relatively inexpensive to
collect a dataset labeled by multiple annotators in a short time. However due to the lack of control over the quality of the annotators, some
abnormal annotators may be affected by position bias which can potentially degrade the quality of the final consensus labels. In this paper we introduce a statistical framework to model and detect annotator's position bias in order to control the false discovery rate (FDR) without a prior knowledge on the amount of biased annotators -- the expected fraction of false discoveries among all discoveries being not too high, in order to assure that most of the discoveries are indeed true and replicable.
The key technical development relies on some new knockoff filters adapted to our problem and new algorithms based on the Inverse Scale Space dynamics whose discretization is potentially suitable for large scale crowdsourcing data analysis.
Our studies are supported by experiments with both simulated examples and real-world data.
The proposed framework provides us a useful tool for quantitatively studying annotator's abnormal behavior in crowdsourcing.

\end{abstract}

\section{Introduction}

In applications, building good
predictive models is challenging primarily due to the difficulties
in obtaining annotated training data.
A traditional way for data labeling
is to hire a small group of experts to provide labels for the entire set of data. However, such an approach can be expensive and time consuming
for large scale data. Thanks to the wide spread of crowdsourcing platforms (e.g.,
\href{https://www.mturk.com}{MTurk},
\href{http://www.innocentive.com/}{Innocentive},
\href{http://crowdflower.com/}{CrowdFlower},
\href{http://www.crowdrank.net/}{CrowdRank}, and
\href{http://www.allourideas.org/}{Allourideas}), a much more efficient way is to post unlabeled data to a crowdsourcing
marketplace, where a big crowd of low-paid workers can be hired instantaneously to perform labeling tasks \cite{Sheng2008,non-expert1,non-expert2,MIR10,MM09}.

Despite of its high efficiency and immediate availability, crowd labeling raises many new challenges. Since
typical crowdsourced tasks are tedious and annotators usually come from a diverse pool including genuine experts,
novices, biased workers, and malicious annotators, labels generated by the crowd suffer from low
quality. Thus, all crowdsourcers need
strategies to ensure the reliability of answers. In other words, outlier detection is a critical task in order to achieve a robust labeling results.
Various methods have been developed in literature for outlier detection, of which majority voting strategy \cite{gygli2013interestingness, jiang2013understanding} is the most typical one. In this setting, each pair is allocated
to multiple annotators and their opinions are averaged over so as to identify and discard noisy
data provided by unreliable raters. They thus require large amount of pairwise labels to be collected.
More importantly as a local outlier detection method, majority voting is ineffective in identifying outliers that can cause global ranking
inconsistencies \cite{fu2014interestingness,fu2015robust}.
The work in~\cite{MM13} attacks this problem
and formulates the outlier detection as a LASSO problem based on sparse approximations of the cyclic ranking
projection of paired comparison data in Hodge decomposition. Regularization paths of the LASSO problem could provide an order on samples tending to be outliers. However, these work all treat pairwise comparison judgements as independent random outliers, which are typically defined to be data samples that have unusual deviations from the remaining data.

In this paper, instead of modeling the random effect of sample-wise outliers, we are primarily interested in the fixed effect where the annotators are influenced by positions when labeling in pairwise comparison setting. 
Such an annotator's position bias \cite{day1969position} is ubiquitous in uncontrolled crowdsourced ranking experiments. In our studies, annotator's position bias typically arises from: i) \textbf{\emph{the ugly}}: one typically clicks one side more often than another. As some pairs are highly confusing or annotators get too tired, in these cases, some annotators tend to click one side hoping to simply raise their record to receive more payment; while for pairs with substantial differences, they click as usual. ii) \textbf{\emph{the bad}}: some extremely careless annotators, or robots pretending to be human annotators, actually do not look at the instances and click one side all the time to quickly receive pay for work.
Such kinds of annotators may significantly deteriorate the quality of crowdsourcing data and increase
the cost of acquiring annotations (since each raw feedback comes with a
cost: the task requestor has to pay workers a pre-specified
monetary reward for each labeling they provide, usually,
regardless of the feedback correctness). 
Although it might be relatively easy to identify the bad annotators above by inspecting their inputs, it is impossible for eye inspection to pick up those ugly annotators with mixed behaviors. Therefore it is desired to design a statistical framework to quantitatively detect and eliminate annotator's position bias
 for crowdsourcing platforms in market. 
Such a systematic study, up to the author's knowledge, however has not been seen in literature.

In this paper, we propose a linear model with annotator's position bias and new algorithms to find good estimates with an automatic control on the false discovery rate (FDR) -- the expected fraction of false discoveries among all discoveries. To understand FDR, imagine that we have a detection method
that has just made 100 discoveries. Then, if our method is known to control the FDR at the 10\%
level, this means that with high probability, we can expect at most 10 of these discoveries to be false and, therefore, at least 90 to be true and replicable. Such a FDR control is desired when we don't have a prior knowledge about the amount of bad or ugly annotators and typical statistical estimates will lead to an over-identification of them. 

Specifically, our contributions in this work are highlighted as follows:

\begin{itemize}
\item[(A)] A linear model with annotator's position bias as fixed effects;
\item[(B)] New algorithms to find good estimates of such position bias, etc., based on Inverse Scale Space dynamics and its discretization Linearized Bregman Iteration;
\item[(C)] New knockoff filters for FDR control adapted to our setting, which aims to mimic the correlation structure found within the original features for position bias;
\item[(D)] Extensive experimental validation based on one simulated and four real-world crowdsourced datasets.
\end{itemize}

\section{Methodology}
In this section, we systematically introduce the methodology for annotator's position bias estimation. Specifically, we first start from a basic linear model with different types of noise models, which have been successfully used widely in literature. Then we introduce a new dynamic approach with unbiased estimator called Inverse Scale Space (ISS). Based on this, we present
the modified knockoff filter for FDR control in details.

\subsection{Basic Linear Model}
Let $V = \{1,2,\dots,n\}$ be the set of nodes and $E = \{(\alpha,i,j): i,j\in V, \alpha \in U\}$ be the set of edges, where $U$ is the set of all annotators. Suppose the pairwise ranking data is given as $Y: E\rightarrow R$. $Y_{ij}^\alpha>0$ means $\alpha$ prefers $i$ to $j$ and $Y_{ij}^{\alpha}\leq 0$ otherwise. The magnitude of $Y_{ij}^\alpha$ can represent the degree of preference and it varies in applications. It can be dichotomous choice $\{\pm 1\}$, $k$-point Likert scale (e.g. $k=3,4,5$), or even real values.

In this paper, consider the following linear model:
\begin{equation} \label{eq:linear}
Y_{ij}^\alpha = \theta_i - \theta_j + z_{ij}^\alpha
\end{equation}
where $\theta: V\to \mathbb{R}$ is some common score on $V$ and the residue $z_{ij}^\alpha$ may have interesting structures in crowdsourcing settings. 

The annotators might have different effects on the residues. While for most annotators, the deviations from the common score are due to random noise; occasionally the annotators deviate from the common behavior regularly -- some careless ones always choose the left or the right candidate in comparisons, but others only do this when they get too confused to decide. Such behaviors can be modeled in the following way,
\begin{equation} \label{eq:bias}
z_{ij}^\alpha=\gamma^\alpha + \varepsilon_{ij}^\alpha,
\end{equation}
where $\gamma^\alpha$ measures an annotator's position bias in a fixed effect, and the remainder $\varepsilon_{ij}^\alpha$ measures the random effect in sampling which is assumed to be sub-gaussian noise. For example, a positive value of $\gamma^\alpha$ means the annotator $\alpha$ is more likely to prefer the left choice. Under the random design of pairwise comparison experiments, a candidate should be placed on the left or the right randomly, so the position should not affect the choice of a careful (good) annotator. Therefore $\gamma^\alpha$ is assumed to be sparse, i.e., zero for most of annotators, and a nonzero position bias $\gamma^\alpha$ means the annotator $\alpha$ is either always choosing one position over the other (bad) or occasionally incurring this when they get too confused or tired (ugly).

We note that \eqref{eq:bias} should not be confused with recent studies in \cite{fu2014interestingness,MM13} on outlier detection problem,
$z_{ij}^\alpha=\gamma_{ij}^\alpha + \varepsilon_{ij}^\alpha$,
where $\gamma_{ij}^\alpha$ models sparse outliers for each sample $(\alpha,i,j)$, which only measures the random effect of samples rather than the fixed effect of annotators. By modeling the annotator's fixed effect on position bias, one can systematically classify the annotators into the good, the ugly, and the bad according to their behaviors.


\subsection{ISS/LBI} \label{sec:ISS}

Define the gradient operator by $\delta_0:\mathbb{R}^{|V|}\to \mathbb{R}^{|E|}$ such that $(\delta_0 \theta)(i,j,\alpha) = \theta_i - \theta_j$, and the annotator operator $A:\R^{|\mathcal{A}|} \to \R^{|E|}$ by $(A\gamma)(i,j,\alpha) = \gamma^\alpha$, then the model above can be rewritten as:
\begin{equation}\label{eq:model1}
Y = \delta_0 \theta + A \gamma +  \varepsilon,
\end{equation}
In this case, detecting the annotators affected by position bias can be reformulated as:
learning a sparse vector $\gamma$ from given data $(\delta_0,A,Y)$. To solve such a problem, in this paper, we choose a new approach based on the following dynamics,
\begin{subequations} \label{eq:iss}
\begin{align}
 \frac{dp}{dt}  &= A^T(Y-\delta_0 \theta-A\gamma)  \label{eq:iss1}\\
 0 &= \delta_0^T (Y-\delta_0 \theta-A\gamma) \label{eq:iss2}\\
 p &\in \partial\|\gamma\|_1. \end{align}
\end{subequations}

Its solution path can be easily solved by a sequence of nonnegative least squares, see \cite{osher2014} and references therein. In this paper we use the free R-package \cite{Libra}.

In \cite{osher2014}, it has been shown that the dynamics above has several advantages over the traditional LASSO approach, which can be formulated as follows in our setting
\begin{equation}\label{eq:lasso}
\min_{\theta,\gamma} \frac{1}{2}\|Y - \delta_0 \theta - A \gamma \|_2^2 + \lambda \|\gamma\|_1.
\end{equation}
First of all, the dynamics above is statistically equivalent to LASSO in terms of model selection consistency but may render oracle estimator which is
bias-free, while the LASSO estimator is well-known biased. In this sense the ISS path can be better than the LASSO path. Here the solution path $\hat{\gamma}(t)_{t:0\rightarrow\infty}$ plays the same role of the regularization path of LASSO $\hat{\gamma}(\lambda)_{\lambda:\infty\rightarrow0}$ with roughly $t=1/\lambda$, where the important features (variables) are selected before the noisy ones. Following the tradition in image processing, such a dynamics is called \emph{Inverse Scale Space} (ISS).

Beyond the charming statistical properties, ISS also admits an extremely simple discrete approximation, i.e., the Linearized Bregman Iteration (LBI), which has been widely used in image reconstruction with TV-regularization. 
Adapted to our setting, the discretized algorithm is illustrated in Algorithm \ref{alg:LB}, which is scalable, easy for parallelization, and particularly suitable for large scale crowdsourced ranking data analysis.
\begin{algorithm}
{{
\caption{LBI in correspondence to (\ref{eq:model1})}\label{alg:LB}
\textbf{Initialization:} Given parameter $\kappa$ and $\triangle t$, define $k=0, w^0=0, \theta^0 = (\delta_0^T\delta_0)^{\dag}\delta_0^TY,\gamma^0=0$.\\
\textbf{Iteration:}
\begin{subequations}
\begin{align}
w^{k+1} &= w^k + A^T(Y - \delta_0\theta^k - A\gamma^k)\triangle t. \label{alg1Step1}\\
\gamma^{k+1}&=\kappa\,\mathrm{shrink}(w^{k+1}). \label{alg1Step2}\\
\theta^{k+1} &= \theta^k + \kappa\delta_0^T (Y-\delta_0 \theta^k-A\gamma^k)\triangle t. \label{alg1Step3}
\end{align}
\end{subequations}
\textbf{Stopping:} exit when $k\triangle t > t $.
}\\
where $\mathrm{shrink}(x) := \mathrm{sign}(x)\max \{|x|-1,0\}$.}
\end{algorithm}

\subsection{FDR Control and New Knockoff Filter} \label{sec:knockoff}
A crucial question for LASSO and ISS is how to choose the regularization parameter $\lambda$ and $t$ in real-world data. After all, different parameters can give different bad or ugly annotator sets. Traditional methods either require a prior knowledge on the amount of such annotators which is often unknown in practice, or some statistically optimal choice of such regularization parameters. Extensive studies in statistics have shown that such parameter tuning typically lead to an over estimation of the sparse signal, therefore False Discovery Rate (FDR) control is necessary  \cite{barber2014controlling} which is adopted in this paper.

FDR is defined as the expected proportion of false discoveries among the discoveries. Putting in a mathematical way, here we consider
\[FDR  = \E \left [\frac{\#\{\alpha:\gamma^\alpha=0,\hat{\gamma}^\alpha\ne 0\}}{\#\{\alpha:\hat{\gamma}^\alpha\ne 0\}\wedge1}\right].\]
To control the FDR means to control the accuracy of the bad/ugly annotators we detected to see if they are reasonable ones.

Recently, a new method called knockoff filter \cite{barber2014controlling} is proposed to automatically control FDR in standard linear regression without a prior knowledge on the sparsity. In this paper, such an approach will be extended to our linear model (\ref{eq:model1}) and the algorithms, where both non-sparse $\theta$ and sparse $\gamma$ co-exist in the model. 
The extended method consists of the same three steps as in  \cite{barber2014controlling}, where the key difference lies in the knockoff feature construction adapted to our setting.

\begin{enumerate}
\item Construct knockoff features: let $\tilde{A}$ be knockoff features that satisfy
\begin{equation}\label{newcond2}
\tilde{A}^T\tilde{A} = A^TA,~ A^T\tilde{A} = A^TA- \diag(s),~\delta_0^T\tilde{A} = \delta_0^TA
\end{equation}
where $s$ is positive and can be solved by SDP:
\begin{eqnarray*}
\max_s   & &\sum_j s_j\\
s.t.	& &0 \le s_j \le 1\\
	& &\diag(s) \preceq 2A^T(I - H)A,
\end{eqnarray*}
with $H:=\delta_0(\delta_0^T\delta_0)^{\dag}\delta_0^T$. Let $Q\in \mathbb{R}^{|E|\times|\mathcal{A}|}$ be an orthonormal matrix such that $\delta_0^TQ=0, A^TQ=0$, which requires $|E|\ge2|\mathcal{A}|+|V|$ easily met in crowdsourcing. Then \eqref{newcond2} can be satisfied by defining
\[\tilde{A} := A - (I - H)A(A^T(I - H)A)^{-1}\diag(s) +QC\]
where $C\in \mathbb{R}^{|\mathcal{A}|\times|\mathcal{A}|}$ satisfies $C^TC = 2\diag(s) - \diag(s)(A^T(I - H)A)^{-1}\diag(s)$.
\\
Now define the extended design matrix $A_{ko} = [A,\tilde{A}]$ and $\gamma_{ko} = [\gamma,\tilde{\gamma}]^T$, then replace $A$ with $A_{ko}$ and $\gamma$ with $\gamma_{ko}$ in \eqref{eq:lasso} , \eqref{eq:iss} or Alg. \ref{alg:LB}, we can get solution path $\hat{\gamma}_{ko}(\lambda)$ (or $\hat{\gamma}_{ko}(t)$).
\item Generate knockoff statistics for every original feature: define $Z_{j}$ to be the first entering time for $A_j$, i.e., $Z_{j} = \sup\{\lambda: \hat{\gamma}_j(\lambda) \neq 0\}$ for LASSO (or $\sup\{1/t: \hat{\gamma}_j(t) \neq 0\}$ for ISS/LBI) and $\tilde{Z}_j$ can be defined similarly. Then the knockoff statistics becomes
\begin{equation}\label{eq:statistic}
W_j = \max(Z_j,\tilde{Z}_j) \sign(Z_j - \tilde{Z}_j)
\end{equation}
\item Choose variables based on the knockoff statistics:
define the selected variable set $\hat{S} = \{j: W_j\ge T_{0/1}\}$, where
\[T_{0/1} = \min \{t: \frac{0/1+\#\{j:W_j\le-t\}}{\#\{j:W_j\ge t\}}\le q\}.\]
$T_0$ is knockoff cut and $T_1$ is knockoff+ cut.
\end{enumerate}



It can be shown that the new knockoff filter above indeed controls FDR in the following sense, whose proof is similar to that of \cite{barber2014controlling} (collected in Supplementary Materials for completeness).

\begin{thm}\label{thm1}
If $\epsilon$ is i.i.d $N(0,\sigma^2)$ and $|E|\ge2|\mathcal{A}|+|V|$, then for any $q\in[0,1]$, the knockoff filter with ISS/LBI (or LASSO) satisfies
\[\E\left[\frac{\#\{j:\gamma_j =0~and~j\in\hat{S}\}}{\#\{j:j\in\hat{S}\}+q^{-1}}\right] \le q\]
and the knockoff+ method satisfies
\[\E\left[\frac{\#\{j:\gamma_j =0~and~j\in\hat{S}\}}{\#\{j:j\in\hat{S}\}}\right] \le q\]
\end{thm}

\begin{remark}\label{rmk1}
There is an equivalent reformulation of (\ref{eq:model1}) to eliminate the non-sparse structure variable $\theta$ and convert it to a standard LASSO. Let $\delta_0$ have a full SVD decomposition $\delta_0 = U \Sigma V^T$ and $U=[U_1,U_2]$, where $U_1$ is an orthonormal basis of the column space $\rm{col}(\delta_0)$ and $U_2$ becomes an orthonormal basis for $\rm{ker} (\delta_0^T)$. Then
\begin{equation}\label{eq:model2}
U_2^TY = U_2^TA \gamma +  U_2^T\varepsilon.
\end{equation}
Let $y = U_2^TY, X = U_2^TA, e = U_2^T\varepsilon$, then $e$ is i.i.d $N(0,\sigma^2)$
\begin{equation}\label{eq:model0}
y = X \gamma +  e.
\end{equation}
Based on this, we can use the original knockoff filter $\tilde{X}$ in \cite{barber2014controlling} to select the position-biased annotators.

A shortcoming of this approach lies in the full SVD decomposition which might be too expensive for large scale problem. The former approach will not suffer from this. However, one can see in the following theorem both approaches are in fact equivalent. Therefore such a reformulation provides us a conceptual insight in understanding the construction of knockoff filters.
\end{remark}

\begin{thm}\label{thm2}
The approach in Remark 1 is equivalent to what we proposed above in the following sense:
\begin{itemize}
\item The knockoff features of \eqref{eq:model0} satisfies $\tilde{X} = U_2^T \tilde{A}$  and $\tilde{A} = U_2\tilde{X} + U_1U_1^TA$;
\item The knockoff statistics constructed by ISS (or LASSO) for both procedures are exactly the same.
\end{itemize}
\end{thm}

Both knockoff filters above can choose variables with FDR control but the estimator $(\hat{\theta},\hat{\gamma}_{ko})$ consists of knockoff features, so we need to reestimate $\hat{\theta},\hat{\gamma}$ after bad annotator detection by passing to a least square while only keeping those nonzero parameters and features. Suppose that $\hat{S}$ is the set of bad or ugly annotators given by knockoff filters, 
then one can find the final estimators by
\begin{equation}
(\hat{\theta},\hat{\gamma}) = \arg\min_{\theta,\gamma_{\hat{S}}} \|Y - \delta_0 \theta - A_{\hat{S}}\gamma_{\hat{S}}\|_2^2.
\end{equation}

 {\renewcommand\baselinestretch{1.3}\selectfont
 \setlength{\belowcaptionskip}{0pt}
\begin{table}\caption{\label{iss_1} Knockoff with $q = 10\%$ via ISS.}
\scriptsize
\centering
\subtable[\small Control of Actual FDR]{
\begin{tabular}{c||ccccc}
  \hline  \textbf{ }   &\textbf{p2=40\%}  &\textbf{p2=50\%} &\textbf{p2=60\%} &\textbf{p2=70\%} \\
 \hline
 \hline  \textbf{p1=10\%}    &0.0959    &0.0833    &0.1198    &0.1229 \\
 \hline  \textbf{p1=20\%}    &0.0917    &0.0989    &0.0935    &0.1006\\
  \hline  \textbf{p1=30\%}   &0.0919    &0.0991    &0.0921    &0.0854 \\
   \hline  \textbf{p1=40\%}   &0.1062    &0.1034    &0.0998    &0.1184  \\

 \hline
 \end {tabular}
 }
\qquad

\subtable[\small Number of True Discoveries]{
\scriptsize
\centering
\begin{tabular}{c||ccccc}
   \hline  \textbf{ }    &\textbf{p2=40\%}  &\textbf{p2=50\%} &\textbf{p2=60\%} &\textbf{p2=70\%} \\
 \hline
 \hline  \textbf{p1=10\%}    &49.95   &50.00   &50.00   &50.00 \\
 \hline  \textbf{p1=20\%}    &49.90   &50.00   &50.00   &50.00 \\
  \hline  \textbf{p1=30\%}  &49.80   &50.00   &50.00   &50.00 \\
   \hline  \textbf{p1=40\%}  &49.75   &49.95   &50.00   &50.00  \\

 \hline
 \end {tabular}
}
\end{table}
 \par}

\section{EXPERIMENTS}\label{sec:experiments}

In this section, five examples are exhibited with both simulated and real-world data to illustrate
the validity of the analysis above and applications of the methodology proposed. The first example is with simulated data while the latter four exploit real-world data collected by crowdsourcing.

\subsection{Simulated Study} \label{sec:simulatedata}

\textbf{Settings} We first validate the proposed algorithm on simulated binary data labeled by 150 annotators. Of the 150 annotators we have 100 \textbf{\emph{good}}
annotators (annotators 1 to 100 without position bias) and 50 \textbf{\emph{bad/ugly}} annotators (annotators 101 to 150 with position bias).
We note that for good annotators, it does not mean
that each worker always present the correct labels. Instead, it means that they also have the probability to make incorrect judgements due to certain reasons, rather than position effect.

Specifically, we first create a random total order on $n$ candidates $V$ as the
ground-truth and add paired comparison edges $(i,j)\in E$ to graph $G=(V,E)$ until a complete graph, with the preference direction following the ground-truth order.
Here we choose $n=|V|=16$, which is consistent with the third real-world dataset with smallest node size. Then, for good annotators, they make judgements with an incorrect probability $p_1$ (i.e., $p_1 \%$ of $E$ is reversed in preference direction), while for bad/ugly annotators, they are with a probability of $p_2$ disturbed by position effect.

\textbf{Evaluation metrics} Two metrics are employed to evaluate the performance of the proposed algorithms. The first one is \emph{Control of Actual FDR}, the second is \emph{Number of True Discoveries}.

\textbf{Experimental results} With different
choices of $p_1$ and $p_2$, the  mean \emph{Control of Actual FDR} and \emph{Number of True Discoveries} with $q = 10\%$ over 100 runs
are shown in Table \ref{iss_1} to measure the performance
of knockoff filter via ISS in position biased annotator detection. It can be seen that via knockoff filter, ISS can provide an accurate detection of position biased annotators
(indicated by \emph{control of actual FDR} around $10\%$ and \emph{Number of True Discoveries} around 50).  Comparable results of LASSO with $q = 10\%$ can be found in Table \ref{lasso_1}. It can
be seen that via knockoff filter, both LASSO and ISS can
provide an accurate detection of position biased annotators.
This result is consistent with the theoretical comparison between LASSO and ISS discussed in \cite{osher2014}, where ISS/LBI has similar theoretical guarantees as LASSO, but with bias-free and simpler implementation (the 3 line algorithm in Sec. \ref{sec:ISS}) properties.

 {\renewcommand\baselinestretch{1.3}\selectfont
 \setlength{\belowcaptionskip}{0pt}
\begin{table}\caption{\label{lasso_1} Knockoff with $q = 10\%$ via LASSO.}
\scriptsize
\centering
\subtable[\small Control of Actual FDR]{
\begin{tabular}{c||ccccc}
 \hline  \textbf{ }    &\textbf{p2=40\%}  &\textbf{p2=50\%} &\textbf{p2=60\%} &\textbf{p2=70\%} \\
 \hline
 \hline  \textbf{p1=10\%}    &0.0711    &0.1326    &0.1433    &0.1256 \\
 \hline  \textbf{p1=20\%}     &0.0998    &0.0954    &0.0970    &0.0780 \\
  \hline  \textbf{p1=30\%}    &0.1044    &0.0918    &0.1093    &0.1061 \\
   \hline  \textbf{p1=40\%}  &0.0843    &0.1035    &0.1063    &0.0941   \\

 \hline
 \end {tabular}
 }
\qquad

\subtable[\small Number of True Discoveries]{
\scriptsize
\centering
\begin{tabular}{c||ccccc}
 \hline  \textbf{ }    &\textbf{p2=40\%}  &\textbf{p2=50\%} &\textbf{p2=60\%} &\textbf{p2=70\%} \\
 \hline
 \hline  \textbf{p1=10\%}   &49.95   &50.00   &50.00   &50.00 \\
 \hline  \textbf{p1=20\%}   &49.90   &50.00   &50.00   &50.00 \\
  \hline  \textbf{p1=30\%}  &49.90   &50.00   &50.00   &50.00 \\
   \hline  \textbf{p1=40\%}  &49.85   &50.00   &50.00   &50.00  \\

 \hline
 \end {tabular}
}
\end{table}
 \par}


\subsection{Real-world Datasets}

As there is no ground-truth for position biased annotators in real-world
data, one can not compute \emph{control of actual FDR} and \emph{Number of True Discoveries} as in simulated
data to evaluate the detection performance
here. In this subsection, we inspect the annotators
returned by knockoff filter via ISS/LASSO under $q = 10\%$ to see if they are reasonably good
position biased workers.

%

{\renewcommand\baselinestretch{1}\selectfont
\setlength{\belowcaptionskip}{3pt}
\begin{table}\caption{\label{age} Position biased annotators detected in Human age dataset, together with the click counts of each side (i.e., Left and Right).}
\tiny
\centering
\newsavebox{\tablebox}
\begin{lrbox}{\tablebox}
\begin{tabular}{||c|c|c||c|c|c||}
  \hline  \textbf{ID}   &\textbf{Left}  &\textbf{Right} & \textbf{ID}   &\textbf{Left}  &\textbf{Right} \\
 \hline
\hline   \textcolor{red}{\textbf{40}}	&40	&0  & \textcolor{blue}{\textbf{50}}	&60	&3 \\
\hline  \textcolor{red}{\textbf{51}}	&63	&0  & \textcolor{blue}{\textbf{59}}	&213	&66 \\
\hline   \textcolor{red}{\textbf{94}}	&0	&30 &    \textcolor{blue}{\textbf{64}}	&5	&14 \\
\hline  \textcolor{blue}{\textbf{12}}	&90	&270  &   \textcolor{blue}{\textbf{70}}	&191	&9 \\
\hline    \textcolor{blue}{\textbf{18}}	&74	&25  &  \textcolor{blue}{\textbf{72}}	&5	&24 \\
\hline   \textcolor{blue}{\textbf{34}}	&32	&48 & \textcolor{blue}{\textbf{77}}	&11	&1 \\
\hline \textcolor{blue}{\textbf{38}}	&110 &15  & \textcolor{blue}{\textbf{81}}	&4	&28 \\
\hline \textcolor{blue}{\textbf{43}} &79	&1    & \textcolor{blue}{\textbf{91}}	&79	&5 \\
\hline \textcolor{blue}{\textbf{46}}	&40	&10   &  \ & \  & \  \\

\hline

\hline
\end{tabular}
\medskip
\end{lrbox}
\scalebox{1.0}{\usebox{\tablebox}}
\end{table}
\par}

\begin{figure}[t]
 \begin{center}
\includegraphics[width=\linewidth]{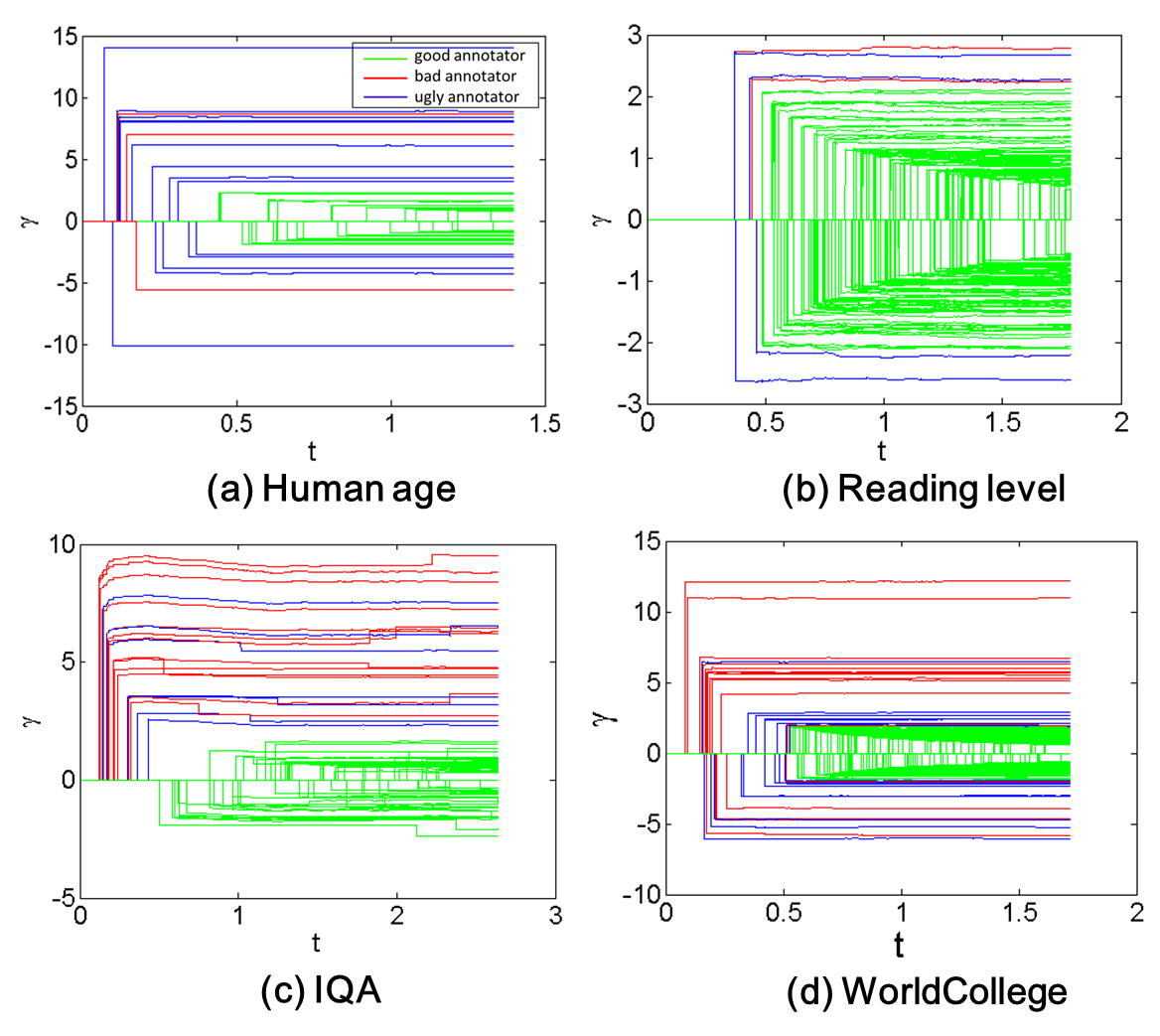}
  \caption{ISS regularization path of four real-world datasets (Green: the good; Red: the bad; Blue: the ugly).} \label{total_path}
\end{center}
\end{figure}

\subsubsection{Human Age}
In this dataset, 30 images from human age dataset FG-NET \footnote{http://www.fgnet.rsunit.com/} are annotated by a group of volunteer users on \href{http://www.chinacrowds.com/}{ChinaCrowds} platform. The groundtruth age ranking is known to us. The annotator is presented with two images and
given a binary choice of which one is older. Totally, we obtain 14,011 pairwise comparisons from 94 annotators.
By adopting the knockoff-based algorithm we proposed, LASSO and ISS identify exactly the same set of abnormal annotators (i.e., 17 users) at q=10\%, as is shown in Table \ref{age}.

It is easy to see that these annotators can be divided into two types: (1) \textbf{\emph{the bad}}: click one side all the time (with ID in red); (2) \textbf{\emph{the ugly}}: click one side with high probability (with ID in blue). Besides, the regularization paths of ISS can be found in Figure \ref{total_path}(a), where the position biased annotators detected mostly lie outside the majority of the paths. Note that since we allow a small percentage of false positives, some ugly annotators might be good in reality as well.

To see the effect of position biased annotators on global ranking scores, Table
\ref{tab:agescore} shows the outcomes of two ranking algorithms, namely
original and corrected. The original is calculated by least squares problems on all of the pairwise comparisons,
while the corrected is obtained by the correction step via knockoff illustrated in Section \ref{sec:knockoff}.
It is easy to see that the removal of position biased annotators often changes the orders of some competitive
images, such as ID=11 and ID=21, ID=30 and ID=8, etc.

To see which ranking is more reasonable, Table \ref{age_gt} shows the \emph{\textbf{groundtruth}} ranking of these competitive images. We can find from this table that, compared with the original ranking, the corrected one is in more agreement with the groundtruth ranking, which further shows that: i) position biased annotators may disturb the ranking to a departure
from the real ranking. ii) pairs with little differences are more likely to lead to position biased annotations.
From this viewpoint, we can see that the knockoff-based FDR-controlling
method indeed effectively selects the position biased annotators.

{\renewcommand\baselinestretch{1.3}\selectfont
\setlength{\belowcaptionskip}{0pt}
\begin{table}
\caption{Comparison of original vs. corrected rankings on Human age dataset. The integer represents the ranking position and the number in parenthesis represents the global ranking score returned by the corresponding algorithm. }\label{tab:agescore}
\centering

\begin{lrbox}{\tablebox}
\tiny

\begin{tabular}{||c|c|c||c|c|c||}
    \hline\hline
    ID & Original.  & Corrected. & ID & Original.  & Corrected. \\
    \hline
 28    & 1 ( 0.7780 )  & 1 ( 0.7573 ) & 23    & 16 ( 0.0208 )    & 16 ( 0.0099 )   \\
 3   & 2 ( 0.6661 )    & 2 ( 0.6771 )  & 8    & 17 ( 0.0086 )    & \textcolor{red}{18 ( -0.0024 )} \\
   14    & 3 ( 0.5653 )   & 3 ( 0.5647 ) & 30   & 18 ( -0.0025 )    & \textcolor{red}{17 ( 0.0055 )} \\
   29   & 4 ( 0.4482 )  & 4 ( 0.4490 ) & 12   & 19 ( -0.0201 )   & 19 ( -0.0632 ) \\
   21    & 5 ( 0.4087 )  & \textcolor{red}{6 ( 0.4086 )} & 13   & 20 ( -0.1961 )   & \textcolor{red}{21 ( -0.2111 )} \\
   11    & 6 ( 0.4059 )   & \textcolor{red}{5 ( 0.4343 )}  & 15   & 21 ( -0.2160 )   & \textcolor{red}{23 ( -0.2791 )} \\
    7    & 7 ( 0.3873 )  & 7 ( 0.4017 ) & 25   & 22 ( -0.2166 )   & \textcolor{red}{20 ( -0.2099 )} \\
    5    & 8 ( 0.3634 )   & 8 ( 0.3478 ) & 16   & 23 ( -0.2551 )   & \textcolor{red}{24 ( -0.2887 )} \\
   27    & 9 ( 0.3582 )     & 9 ( 0.3377 )  & 2   & 24 ( -0.3710 )    & \textcolor{red}{22 ( -0.2785 )} \\
   24    & 10 ( 0.2064 )    & 10 ( 0.1722 )  & 9   & 25 ( -0.4158 )    & 25 ( -0.3949 ) \\
    6    & 11 ( 0.0932 )      & \textcolor{red}{13 ( 0.1084 )} & 1   & 26 ( -0.6135 )   & \textcolor{red}{27 ( -0.6376 )} \\
    4    & 12 ( 0.0914 )        & \textcolor{red}{12 ( 0.1207 )} & 18   & 27 ( -0.6249 )  & \textcolor{red}{26 ( -0.6180 )} \\
   22    & 13 ( 0.0896 )     & \textcolor{red}{11 ( 0.1032 )}  & 19   & 28 ( -0.6653 )   & 28 (  -0.6390 ) \\
   17    & 14 ( 0.0872 )  & 14 ( 0.1232 ) & 10   & 29 ( -0.6969 )   & 29 ( -0.7040 ) \\
   20    & 15 ( 0.0816 )    & 15 ( 0.0559 )  & 26   & 30 ( -0.7660 )    & 30 ( -0.7509 ) \\

   \hline
   \hline
\end{tabular}
\medskip
\end{lrbox}
\scalebox{0.9}{\usebox{\tablebox}}
\end{table}
\par}

{\renewcommand\baselinestretch{1.3}\selectfont
\setlength{\belowcaptionskip}{0pt}
\begin{table}\caption{\label{age_gt} Groundtruth ranking of the competitive images highlighted with red color in Table \ref{tab:agescore}.}

\centering
\tiny
\begin{lrbox}{\tablebox}
\begin{tabular}{||c||}
 \hline
 \hline   $11 \succ 21$ \\
  \hline   $22 \succ 4 \succ 6$ \\
   \hline   $30 \succ \succ 8$ \\
   \hline    $25 \succ 13 \succ 16 \succ 2 \succ 15$ \\
   \hline   $18 \succ 1$ \\
\hline
 \hline
 \end {tabular}
\medskip
\end{lrbox}
\scalebox{1.0}{\usebox{\tablebox}}
\end{table}
\par}

\subsubsection{Reading Level}

The second dataset is a subset of reading level dataset \cite{chen2013pairwise}, which contains 490 documents. 8,000 pairwise
comparisons are collected from 346 annotators using \href{http://www.crowdflower.com/}{CrowdFlower} crowdsourcing platform. More specifically, each
annotator is asked to provide his/her opinion on which text is more challenging to read and understand.
 Table \ref{readinglevel} shows the position biased annotators detected from this dataset, together with the ISS regularization path shown in Figure \ref{total_path}(b). It is easy to see that LASSO and ISS picked out the same 6 annotators as position biased ones. In terms of the small number of bad annotators detected, we can say that the overall quality of
annotators on this task is relatively high.

{\renewcommand\baselinestretch{1}\selectfont
\setlength{\belowcaptionskip}{0pt}
\begin{table}\caption{\label{readinglevel} Position biased annotators detected in Reading level.}
\tiny
\centering
\begin{lrbox}{\tablebox}
\begin{tabular}{||c|c|c||}

  \hline  \textbf{ID}   &\textbf{Left}  &\textbf{Right} \\
 \hline

\hline   \textcolor{red}{\textbf{50}} &5	 &0 \\
\hline   \textcolor{red}{\textbf{69}} &6	&0 \\
\hline   \textcolor{blue}{\textbf{122}} &19	&3 \\
\hline   \textcolor{blue}{\textbf{148}} &4	&19 \\
\hline   \textcolor{blue}{\textbf{167}} &22	&8 \\
\hline   \textcolor{blue}{\textbf{275}} &7	&22 \\
 \hline
\end {tabular}
\medskip
\end{lrbox}
\scalebox{1.0}{\usebox{\tablebox}}
\end{table}
\par}

\subsubsection{Image Quality Assessment}

%
%

%

%
%

The third dataset is a pairwise comparison dataset for subjective image quality assessment (IQA), which contains 15 reference images and 15 distorted versions of each reference, for a total of 240 images which come from two publicly available datasets LIVE, \cite{LIVE} and IVC \cite{IVC}. Totally, 342 observers,
each of whom performs a varied number of comparisons via Internet, provide
$52,043$ paired comparisons for crowdsourced subjective image quality assessment. Note that the number of responses each reference
image received is different in this dataset.

To validate whether the annotators we detected are good position biased annotators or not, we randomly take
reference image 1 as an illustrative example while other reference
images exhibit similar results. Table \ref{ref1} shows the annotators with position bias picked by knockoff filter and the ISS regularization path is shown in Figure \ref{total_path}(c). In this dataset, the abnormal annotators picked out by LASSO and ISS are also exactly the same. It is easy to see that annotators picked out
are mainly those clicking on one side almost all the time. Besides, it is interesting to see that all these bad annotators highlighted with
red color in Table \ref{ref1} click the left side all the time. We then go back to the crowdsourcing platform and find out that the reason behind this is a default choice on the left button
thus induces some lazy annotators cheat for the annotation task.

{\renewcommand\baselinestretch{1}\selectfont
\setlength{\belowcaptionskip}{0pt}
\begin{table}\caption{\label{ref1} Position biased annotators detected in reference image 1.}
\tiny
\centering
\begin{lrbox}{\tablebox}
\begin{tabular}{||c|c|c||c|c|c||}
  \hline  \textbf{ID}   &\textbf{Left}  &\textbf{Right} & \textbf{ID}   &\textbf{Left}  &\textbf{Right} \\
 \hline
 \hline  \textcolor{red}{\textbf{2}}   &55     &0   & \textcolor{red}{\textbf{300}}   &11     &0\\
 \hline  \textcolor{red}{\textbf{23}}    &42     &0  & \textcolor{red}{\textbf{317}}    &20     &0\\
  \hline  \textcolor{red}{\textbf{29}}  &58     &0  & \textcolor{red}{\textbf{334}}     &90    & 0\\
  \hline  \textcolor{red}{\textbf{99}}   &29     &0 & \textcolor{blue}{\textbf{33}}   &15     &1 \\
  \hline  \textcolor{red}{\textbf{177}}   &77     &0 & \textcolor{blue}{\textbf{34}}   & 8     &1  \\
\hline  \textcolor{red}{\textbf{190}}   &36     &0 &  \textcolor{blue}{\textbf{103}}  &74     &4  \\
\hline  \textcolor{red}{\textbf{228}}   &14     &0  &  \textcolor{blue}{\textbf{133}}  &20    &11\\
\hline  \textcolor{red}{\textbf{241}}    &22     &0 &  \textcolor{blue}{\textbf{207}}    &46     &2\\
\hline  \textcolor{red}{\textbf{259}}    &96     &0 &  \textcolor{blue}{\textbf{260}}    & 49     &2\\
\hline  \textcolor{red}{\textbf{287}}   &34     &0 &  \textcolor{blue}{\textbf{304}}   &17     &1\\

 \hline
\end {tabular}
\medskip
\end{lrbox}
\scalebox{1.0}{\usebox{\tablebox}}
\end{table}
\par}

\subsubsection{WorldCollege Ranking}

We now apply the knockoff filter to the WorldCollege dataset, which is composed of 261 colleges. Using the \href{http://www.allourideas.org/}{Allourideas} crowdsourcing platform, a total of
340 distinct annotators from various countries (e.g., USA, Canada, Spain, France, Japan)
are shown randomly with
pairs of these colleges, and asked to decide which of the
two universities is more attractive to attend. Finally, we obtain a
total of 8,823 pairwise comparisons. We then apply knockoff filter to the resulting dataset and find out that both LASSO and ISS selected 36 annotators as position biased ones, as is shown in Table \ref{college} and Figure \ref{total_path}(d). It is easy to see that similar to the human age dataset, the annotators picked out are either clicking one side all the time, or
clicking one side with high probability.

%

{\renewcommand\baselinestretch{1}\selectfont
\setlength{\belowcaptionskip}{0pt}
\begin{table}\caption{\label{college} Position biased annotators detected in WorldCollege.}
\tiny
\centering
\begin{lrbox}{\tablebox}
\begin{tabular}{||c|c|c||c|c|c||}
  \hline  \textbf{ID}   &\textbf{Left}  &\textbf{Right} & \textbf{ID}   &\textbf{Left}  &\textbf{Right} \\
 \hline

 \hline  \textcolor{red}{\textbf{56}}	    &17	&0 & \textcolor{blue}{\textbf{25}}	&17	&6 \\
 \hline  \textcolor{red}{\textbf{75}}	    &0	&3  & \textcolor{blue}{\textbf{59}}	&9	&29 \\
 \hline  \textcolor{red}{\textbf{101}}	&26	&0  & \textcolor{blue}{\textbf{87}}	&11	&62 \\
 \hline  \textcolor{red}{\textbf{115}}	&34	&0  & \textcolor{blue}{\textbf{122}}	&13	&9 \\
 \hline  \textcolor{red}{\textbf{145}}	&0	 &27  & \textcolor{blue}{\textbf{134}}	&20	&7 \\
 \hline  \textcolor{red}{\textbf{166}}	&35	&0 & \textcolor{blue}{\textbf{140}}	&12	&4 \\
 \hline  \textcolor{red}{\textbf{209}}	&127 &0  & \textcolor{blue}{\textbf{156}}	&189	&67 \\
 \hline  \textcolor{red}{\textbf{222}}	&0	&2  & \textcolor{blue}{\textbf{189}}	&2	&12 \\
 \hline  \textcolor{red}{\textbf{245}}	&0	&34 & \textcolor{blue}{\textbf{191}}	&23	&7 \\
 \hline  \textcolor{red}{\textbf{256}}	&0	&21  & \textcolor{blue}{\textbf{202}}	&2	&8 \\
 \hline  \textcolor{red}{\textbf{267}}	&45	&0  & \textcolor{blue}{\textbf{207}}	&23	&10 \\
 \hline  \textcolor{red}{\textbf{268}} &148	&0  & \textcolor{blue}{\textbf{208}}	&10	&2 \\
 \hline  \textcolor{red}{\textbf{275}} &1	&0  & \textcolor{blue}{\textbf{239}}	&11	&2 \\
 \hline  \textcolor{red}{\textbf{289}} &35	&0  & \textcolor{blue}{\textbf{258}}	&2	&13 \\
 \hline  \textcolor{red}{\textbf{299}}	&31	&0   & \textcolor{blue}{\textbf{270}}	&20	&70 \\
 \hline \textcolor{red}{\textbf{321}}	&33	&0  & \textcolor{blue}{\textbf{276}}	&16	&54 \\
 \hline \textcolor{red}{\textbf{323}}	&35	&0 & \textcolor{blue}{\textbf{320}}	&253	&324 \\
 \hline \textcolor{red}{\textbf{338}}	&0	&21  & \textcolor{blue}{\textbf{330}}	&4	&10 \\

 \hline
\end {tabular}
\medskip
\end{lrbox}
\scalebox{1}{\usebox{\tablebox}}
\end{table}
\par}

\subsection{Discussion}

\begin{figure}[t]
 \begin{center}
\includegraphics[width=\linewidth]{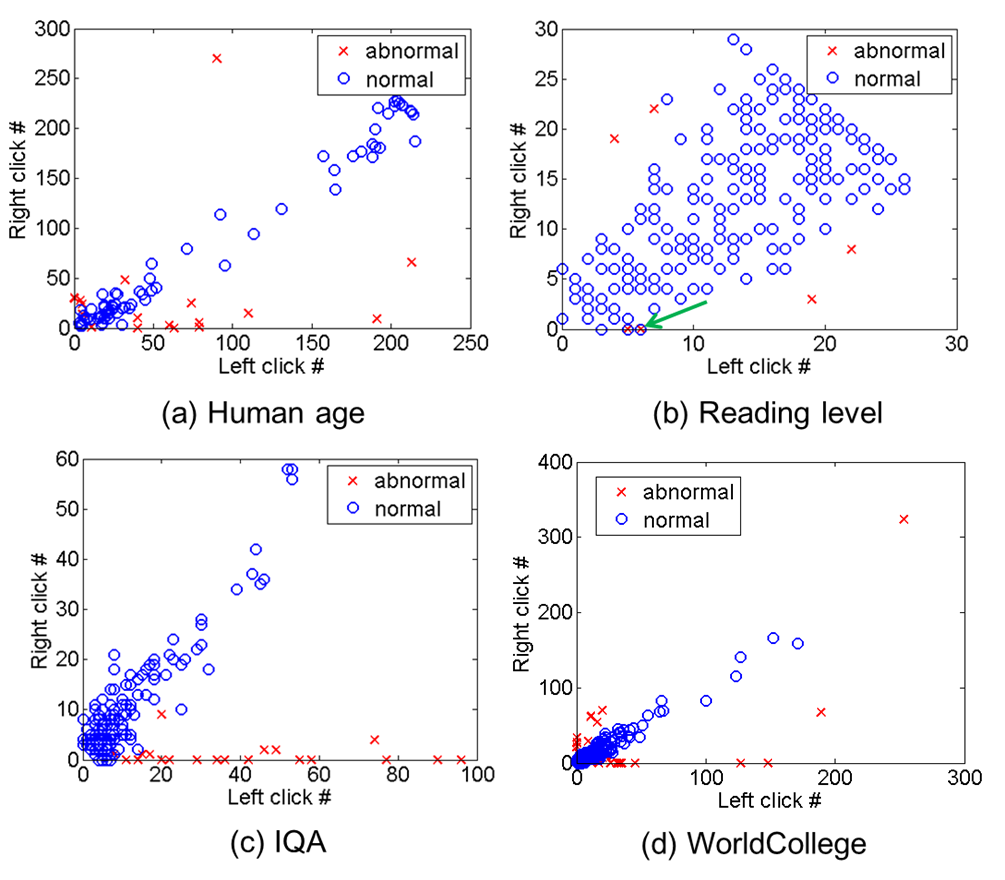}
  \caption{Number of left clicks vs. right clicks of abnormal and normal annotators on four real-world datasets.} \label{leftvsright}
\end{center}
\end{figure}


Someone may argue that setting a threshold on the ratio of left/right answers can be an easy way to detect position biased annotators. To illustrate why simply setting a threshold does not work, Figure \ref{leftvsright} shows the click counts of each side (i.e., X-axis: number of left clicks; Y-axis: number of right clicks), where each
color $\circ/\times$ represents one annotator. It is easy to see that there are indeed some overlaps between abnormal and normal annotators. For example, in reading level dataset, annotators with ID=69 and ID=57 both provide 6:0 on the ratio of left/right clicks. However, ID=69 is detected as abnormal annotator, while ID=57 as normal one. To figure out the reason behind this, we further compute the Match Ratio (MR) of these two annotators with the global ranking scores obtained by all pairwise comparisons and find that $MR_{ID=69}=3/6$ and $MR_{ID=57}=5/6$. This indicates that the position biased annotator (i.e., ID=69) we picked out is the one not only with one-side click but also with a large deviation with the majority. Similar results can be easily found in other three datasets.

\section{Related Work}

%
%
%
%
%
\subsection{Outlier Detection}

Outliers are often referred to as abnormalities, discordants, deviants,
or anomalies in data. Generally speaking, there can be two types of outliers: (1) samples as outliers; (2) subjects as outliers. Hawkins formally
defined in~\cite{hawkins1980identification} the concept of an outlier as follows: ``An outlier is an observation
which deviates so much from the other observations as to arouse suspicions that it was generated by a different
mechanism." Outliers are rare events, but once they have occurred, they may lead to a large instability of models
estimated from the noisy data. For type (1), many methods have been developed for outlier detection, such as distribution-based~\cite{barnett1994outliers}, depth-based~\cite{johnson1998fast}, distance-based~\cite{knorr1999finding,knorr2000distance},
density-based~\cite{breunig2000lof}, and clustering-based~\cite{jain1999data} methods. For subject-based outlier detection, some sophisticated methods have been proposed to model annotators' judgements.
Recently, \cite{chen2013pairwise} propose a Crowd-BT algorithm to detect spammers and malicious annotators: spammers assign random
labels while malicious annotators assign the wrong label most of the time. Besides, \cite{raykar2011ranking} defines
a score to rank the annotators for crowdsourced labeling tasks. Furthermore,
\cite{raykar2012} presents an empirical Bayesian algorithm called SpEM to eliminate the spammers and
estimate the consensus labels based only on the good annotators. However, a phenomenon that has annoyed researchers who
have used paired comparison tests is position bias or
testing order effects. Until now, little work have been found for such kind of position biased annotator detection, which is our main focus in this paper.

\subsection{FDR Control and Knockoff Method}
Most variable selection techniques in statistics such as LASSO suffer from over-selection as picking up too many false positives by leaving out few true positives. In order to offer guarantees on the accuracy of the selection, it is desired to control the false discovery rate (FDR) among all the selected variables. The Benjamini-Hochberg (BH) procedure \cite{benjamini1995controlling} is a typical method known to control FDR under independence scenarios. Recently, \cite{barber2014controlling} developed a new knockoff filter filtermethod for FDR control for general dependent features as long as the sample size is larger than that of parameters. In this paper, we extend this method to our setting with mixed parameters of both nonsparse and sparse ones to achieve the same FDR control.

\subsection{Inverse Scale Space and Linearized Bregman Iteration}

Linearized Bregman Iteration (LBI) has been widely used in image processing and compressed sensing \cite{OBG+05,YODG08} even before its limit form as Inverse Scale Space (ISS) dynamics \cite{burger2005nonlinear}. ISS/LBI at least have two advantages over the popular LASSO in variable selection: (1) ISS may give unbiased estimator \cite{osher2014}, under nearly the same condition for model selection consistency as LASSO whose estimators are however always biased \cite{Fan2001}. (2) LBI, regarded as a discretization of ISS dynamics, is an extremely simple algorithm which combines an iterative gradient descent algorithm together with a soft thresholding. It only runs in a single path and regularization is achieved by early stopping like boosting algorithms \cite{osher2014}, which may save the computational cost greatly and thus suitable for large scale implementation \cite{LBI_decentral}.

\section{Conclusion}

Annotator's position bias is ubiquitous in crowdsourced ranking data, which, up to our knowledge, has not been systematically addressed in literature. In this paper, we propose a statistical model for annotator's position bias with pairwise comparison data on graphs, together with new algorithms to reach statistically good estimates with a FDR control based on some new design of knockoff filters. FDR control here does not need a prior knowledge on the sparsity of position bias, i.e., the amount of bad or ugly annotators. Such a framework is valid for both traditional LASSO estimator and the new dynamic approach based on ISS/LBI with debiased estimator and scalable implementations which is desired for crowdsourcing experiments. Experimental studies are conducted with both simulated examples and real-world datasets. Our results suggest that the proposed methodology is an effective tool to investigate annotator's abnormal behavior in modern crowdsourcing data.

\section*{Acknowledgements}

The research of Qianqian Xu was supported in part by National Natural Science Foundation of China (No. 61422213, 61402019, 61390514, 61572042), China Postdoctoral Science Foundation (2015T80025), ``Strategic Priority Research Program" of the Chinese Academy of Sciences (XDA06010701), and National Program for Support of Top-notch Young Professionals.
The research of Jiechao Xiong and Yuan Yao
was supported in part by National Basic Research Program of China under
grant 2015CB85600, 2012CB825501, and NSFC grant 61370004, 11421110001
(A3 project), as well as grants from Baidu and Microsoft Research-Asia. Xiaochun Cao and Yuan Yao are the corresponding authors. We would like to
thank Yongyi Guo for helpful discussions and anonymous reviewers who gave
valuable suggestions to help improve the manuscript.

\bibliography{egbib}

\begin{thebibliography}{32}
\providecommand{\natexlab}[1]{#1}
\providecommand{\url}[1]{\texttt{#1}}
\expandafter\ifx\csname urlstyle\endcsname\relax
  \providecommand{\doi}[1]{doi: #1}\else
  \providecommand{\doi}{doi: \begingroup \urlstyle{rm}\Url}\fi

\bibitem[IVC(2005)]{IVC}
Subjective quality assessment irccyn/ivc database.
\newblock \url{http://www2.irccyn.ec-nantes.fr/ivcdb/}, 2005.

\bibitem[LIV(2008)]{LIVE}
{L}{I}{V}{E} image \& video quality assessment database.
\newblock \url{http://live.ece.utexas.edu/research/quality/}, 2008.

\bibitem[Barber \& Cand\`{e}s(2015)Barber and
  Cand\`{e}s]{barber2014controlling}
Barber, R. and Cand\`{e}s, E.
\newblock Controlling the false discovery rate via knockoffs.
\newblock \emph{The Annals of Statistics}, 43\penalty0 (5):\penalty0
  2055--2085, 2015.

\bibitem[Barnett \& Lewis(1994)Barnett and Lewis]{barnett1994outliers}
Barnett, V. and Lewis, T.
\newblock \emph{Outliers in statistical data}, volume~3.
\newblock Wiley New York, 1994.

\bibitem[Benjamini \& Hochberg(1995)Benjamini and
  Hochberg]{benjamini1995controlling}
Benjamini, Y. and Hochberg, Y.
\newblock Controlling the false discovery rate: a practical and powerful
  approach to multiple testing.
\newblock \emph{Journal of the Royal Statistical Society. Series B
  (Methodological)}, 57\penalty0 (1):\penalty0 289--300, 1995.

\bibitem[Breunig et~al.(2000)Breunig, Kriegel, Ng, and Sander]{breunig2000lof}
Breunig, M., Kriegel, H., Ng, R., and Sander, J.
\newblock L{O}{F}: identifying density-based local outliers.
\newblock In \emph{Proceedings of the ACM International Conference on
  Management of Data}, volume~29, pp.\  93--104, 2000.

\bibitem[Burger et~al.(2005)Burger, Osher, Xu, and Gilboa]{burger2005nonlinear}
Burger, M., Osher, S., Xu, J., and Gilboa, G.
\newblock \emph{Nonlinear inverse scale space methods for image restoration}.
\newblock Springer, 2005.

\bibitem[Chen et~al.(2009)Chen, Wu, Chang, and Lei]{MM09}
Chen, K., Wu, C., Chang, Y., and Lei, C.
\newblock A crowdsourceable {Q}o{E} evaluation framework for multimedia
  content.
\newblock In \emph{ACM International Conference on Multimedia}, pp.\  491--500,
  2009.

\bibitem[Chen et~al.(2013)Chen, Bennett, Collins-Thompson, and
  Horvitz]{chen2013pairwise}
Chen, X., Bennett, P., Collins-Thompson, K., and Horvitz, E.
\newblock Pairwise ranking aggregation in a crowdsourced setting.
\newblock In \emph{International Conference on Web Search and Data Mining},
  pp.\  193--202, 2013.

\bibitem[Day(1969)]{day1969position}
Day, R.
\newblock Position bias in paired product tests.
\newblock \emph{Journal of Marketing Research}, 6\penalty0 (1):\penalty0
  98--100, 1969.

\bibitem[Fan \& L(2001)Fan and L]{Fan2001}
Fan, J. and L, R.
\newblock Variable selection via nonconcave penalized likelihood and its oracle
  properties.
\newblock \emph{Journal of American Statistical Association}, 96\penalty0
  (456):\penalty0 1348--1360, 2001.

\bibitem[Fu et~al.(2014)Fu, Hospedales, Xiang, Gong, and
  Yao]{fu2014interestingness}
Fu, Y., Hospedales, T., Xiang, T., Gong, S., and Yao, Y.
\newblock Interestingness prediction by robust learning to rank.
\newblock In \emph{European Conference on Computer Vision}, pp.\  488--503.
  2014.

\bibitem[Fu et~al.(2016)Fu, Hospedales, Xiang, Xiong, Gong, Wang, and
  Yao]{fu2015robust}
Fu, Y., Hospedales, T., Xiang, T., Xiong, J., Gong, S., Wang, Y., and Yao, Y.
\newblock Robust subjective visual property prediction from crowdsourced
  pairwise labels.
\newblock \emph{IEEE Transactions on Pattern Analysis and Machine
  Intelligence}, 38\penalty0 (3):\penalty0 563--577, 2016.

\bibitem[Gygli et~al.(2013)Gygli, Grabner, Riemenschneider, Nater, and
  Gool]{gygli2013interestingness}
Gygli, M., Grabner, H., Riemenschneider, H., Nater, F., and Gool, L.
\newblock The interestingness of images.
\newblock In \emph{IEEE International Conference on Computer Vision}, pp.\
  1633--1640, 2013.

\bibitem[Hawkins(1980)]{hawkins1980identification}
Hawkins, D.
\newblock \emph{Identification of Outliers}, volume~11.
\newblock Springer, 1980.

\bibitem[Hsueh et~al.(2009)Hsueh, Melville, and Sindhwani]{non-expert2}
Hsueh, P., Melville, P., and Sindhwani, V.
\newblock Data quality from crowdsourcing: a study of annotation selection
  criteria.
\newblock In \emph{NAACL HLT Workshop on Active Learning for Natural Language
  Processing}, pp.\  27--35, 2009.

\bibitem[Jain et~al.(1999)Jain, Murty, and Flynn]{jain1999data}
Jain, A., Murty, M., and Flynn, P.
\newblock Data clustering: a review.
\newblock \emph{ACM Computing Surveys}, 31\penalty0 (3):\penalty0 264--323,
  1999.

\bibitem[Jiang et~al.(2013)Jiang, Wang, Feng, Xue, Zheng, and
  Yang]{jiang2013understanding}
Jiang, Y., Wang, Y., Feng, R., Xue, X., Zheng, Y., and Yang, H.
\newblock Understanding and predicting interestingness of videos.
\newblock In \emph{AAAI Conference on Artificial Intelligence}, volume~1, pp.\
  ~2, 2013.

\bibitem[Johnson et~al.(1998)Johnson, Kwok, and Ng]{johnson1998fast}
Johnson, T., Kwok, I., and Ng, R.
\newblock Fast computation of 2-dimensional depth contours.
\newblock In \emph{ACM International Conference on Knowledge Discovery and Data
  Mining}, pp.\  224--228, 1998.

\bibitem[Knorr \& Ng(1999)Knorr and Ng]{knorr1999finding}
Knorr, E. and Ng, R.
\newblock Finding intensional knowledge of distance-based outliers.
\newblock In \emph{International Conference on Very Large Data Bases}, pp.\
  211--222, 1999.

\bibitem[Knorr et~al.(2000)Knorr, Ng, and Tucakov]{knorr2000distance}
Knorr, E., Ng, R., and Tucakov, V.
\newblock Distance-based outliers: algorithms and applications.
\newblock \emph{International Journal on Very Large Data Bases}, 8\penalty0
  (3-4):\penalty0 237--253, 2000.

\bibitem[Nowak \& R{\"u}ger(2010)Nowak and R{\"u}ger]{MIR10}
Nowak, S. and R{\"u}ger, S.
\newblock How reliable are annotations via crowdsourcing: a study about
  inter-annotator agreement for multi-label image annotation.
\newblock In \emph{International Conference on Multimedia Information
  Retrieval}, pp.\  557--566, 2010.

\bibitem[Osher \& Yin(2005)Osher and Yin]{OBG+05}
Osher, S., Burger M. Goldfarb D. Xu~J. and Yin, W.
\newblock An iterative regularization method for total variation-based image
  restoration.
\newblock \emph{SIAM Journal on Multiscale Modeling and Simulation}, 4\penalty0
  (2):\penalty0 460--489, 2005.

\bibitem[Osher et~al.(2016)Osher, Ruan, Xiong, Yao, and Yin]{osher2014}
Osher, S., Ruan, F., Xiong, J., Yao, Y., and Yin, W.
\newblock Sparse recovery via differential inclusions.
\newblock \emph{Applied and Computational Harmonic Analysis}, 2016.
\newblock \doi{10.1016/j.acha.2016.01.002}.

\bibitem[Raykar \& Yu(2011)Raykar and Yu]{raykar2011ranking}
Raykar, V. and Yu, S.
\newblock Ranking annotators for crowdsourced labeling tasks.
\newblock In \emph{Advances in Neural Information Processing Systems}, pp.\
  1809--1817, 2011.

\bibitem[Raykar \& Yu(2012)Raykar and Yu]{raykar2012}
Raykar, V. and Yu, S.
\newblock Eliminating spammers and ranking annotators for crowdsourced labeling
  tasks.
\newblock \emph{The Journal of Machine Learning Research}, 13\penalty0
  (1):\penalty0 491--518, 2012.

\bibitem[Sheng et~al.(2008)Sheng, Provost, and Ipeirotis]{Sheng2008}
Sheng, V., Provost, F., and Ipeirotis, P.
\newblock Get another label? improving data quality and data mining using
  multiple, noisy labelers.
\newblock In \emph{ACM International Conference on Knowledge Discovery and Data
  Mining}, pp.\  614--622, 2008.

\bibitem[Snow et~al.(2008)Snow, O'Connor, Jurafsky, and Ng]{non-expert1}
Snow, R., O'Connor, B., Jurafsky, D., and Ng, A.
\newblock Cheap and fast|but is it good?: evaluating non-expert annotations for
  natural language tasks.
\newblock In \emph{Conference on Empirical Methods in Natural Language
  Processing}, pp.\  254--263, 2008.

\bibitem[Xiong et~al.(2016)Xiong, Ruan, and Yao]{Libra}
Xiong, J., Ruan, F., and Yao, Y.
\newblock \emph{A Tutorial on Libra: R package for the Linearized Bregman
  Algorithm in high dimensional statistics}, 2016.
\newblock URL \url{https://cran.r-project.org/web/packages/Libra}.
\newblock arXiv:1604.05910.

\bibitem[Xu et~al.(2013)Xu, Xiong, Huang, and Yao]{MM13}
Xu, Q., Xiong, J., Huang, Q., and Yao, Y.
\newblock Robust evaluation for quality of experience in crowdsourcing.
\newblock In \emph{ACM International Conference on Multimedia}, pp.\  43--52,
  2013.

\bibitem[Yin et~al.(2008)Yin, Osher, D., and G.]{YODG08}
Yin, W., Osher, S., D., Jerome, and G., Donald.
\newblock Bregman iterative algorithms for compressed sensing and related
  problems.
\newblock \emph{SIAM Journal on Imaging Sciences}, 1\penalty0 (1):\penalty0
  143--168, 2008.

\bibitem[Yuan et~al.(2013)Yuan, Ling, Yin, and Ribeiro]{LBI_decentral}
Yuan, K., Ling, Q., Yin, W., and Ribeiro, A.
\newblock {A Linearized Bregman Algorithm for Decentralized Basis Pursuit}.
\newblock \emph{European Signal Processing Conference}, pp.\  1--5, 2013.

\end{thebibliography}
\bibliographystyle{icml2016}

\clearpage

\setcounter{page}{1}
\section*{Supplementary Material}
{\bf{Sketchy Proof of Theorem \ref{thm1}.}}

Similar to the treatment in \cite{barber2014controlling}, we only need to prove that the knockoff statistics $W_j$ satisfy the following two properties:
\begin{itemize}
\item \emph{sufficiency property}: \\
$W = f([\delta_0,A_{ko}]^T[\delta_0,A_{ko}], [\delta_0,A_{ko}]^TY)$, which indicates $W$ depends only on $[\delta_0,A_{ko}]^T[\delta_0,A_{ko}]$ and $[\delta_0,A_{ko}]^TY$.
\item \emph{antisymmetry property}:\\
 Swapping $A_j$ and $\tilde{A}_j$ has the effect of switching the sign of $W_j$.
\end{itemize}
The second property is obvious because $W_j$ is constructed  using entering time difference. Now we go to prove the first property.

For ISS and LBI, the whole path is only determined by
\begin{align*}
 A_{ko}^T(Y-\delta_0 \theta-A_{ko}\gamma_{ko}) &=  A_{ko}^TY -  A_{ko}^T[\delta_0 ,A_{ko}][\theta^T,\gamma_{ko}^T]^T),  \\
\delta_0^T (Y-\delta_0 \theta-A_{ko}\gamma_{ko}) &=  \delta_0^TY -  \delta_0^T[\delta_0 ,A_{ko}][\theta^T,\gamma_{ko}^T]^T),
\end{align*}
which is only based on $[\delta_0,A_{ko}]^T[\delta_0,A_{ko}]$ and $[\delta_0,A_{ko}]^TY$, so is the entering time $Z_{j}$

The same reasoning holds for LASSO since
\begin{equation*}
\min_{\theta,\gamma} \frac{1}{2}\|Y - [\delta_0,A_{ko}] [\theta^T,\gamma_{ko}^T]^T\|_2^2 + \lambda \|\gamma_{ko}\|_1
\end{equation*}
is equivalent to
\begin{eqnarray*}
\min_{\theta,\gamma} &\frac{1}{2}(\|Y\|_2^2 + [\theta^T,\gamma_{ko}^T] [\delta_0,A_{ko}] ^T[\delta_0,A_{ko}] [\theta^T,\gamma_{ko}^T]^T\\
	&-2 [\theta^T,\gamma_{ko}^T] [\delta_0,A_{ko}] ^TY )+ \lambda \|\gamma_{ko}\|_1
\end{eqnarray*}
So the entire path is determined by $[\delta_0,A_{ko}]^T[\delta_0,A_{ko}]$ and $[\delta_0,A_{ko}]^TY$.
\\
\\

{\bf{Proof of Theorem \ref{thm2}.}}

Suppose $\tilde{X}$ is the knockoff statistics for \eqref{eq:model0}, then it satisfies
\begin{equation}\label{cond2}
\tilde{X}^T\tilde{X} = X^TX, X^T\tilde{X} = X^TX - \diag(s).
\end{equation}
Let $B = A + U_2(\tilde{X} - X)$, then $\tilde{X} = U_2^TB$ and it can be verified
\begin{equation*}
B^TB= A^TA, A^TB = A^TA-\diag(s), \delta_0^TB=\delta_0^TA
\end{equation*}
which means $B$ is a valid knockoff feature matrix for \eqref{eq:model1}.

On the reverse, let $\tilde{A}$ be knockoff features for \eqref{eq:model1}, it is also easy to verify $\tilde{X} = U_2^T\tilde{A}$ satisfies condition \eqref{cond2}. This establishes an injection between $\tilde{X}$ and  $\tilde{A}$.

The equivalence of knockoff statistics comes from the equivalence of solution paths in both approaches. To see this, \eqref{eq:iss2} actually means $\hat{\theta} = (\delta_0^T\delta_0)^{\dag}\delta_0^T(Y - A_{ko}\gamma_{ko})$, plugging $\hat{\theta}$ in \eqref{eq:iss1}, we get
 \begin{eqnarray*}
 \frac{dp}{dt} &=&A_{ko}^T (Y-\delta_0 \hat{\theta}-A_{ko}\gamma_{ko}) \\
&=& A_{ko}^T(U_2U_2^T(Y-A_{ko}\gamma_{ko}))\\
 &= & (U_2^TA_{ko})^T(U_2^TY-U_2^TA_{ko}\gamma_{ko})
 \end{eqnarray*}
This is equivalent to the ISS for the second procedure model \eqref{eq:model2} in Remark 1. So in both approaches, the two ISS solution paths are identical.

The same reasoning holds for LASSO, the derivative of \eqref{eq:lasso} w.r.t. $\theta$ is zero at the optimal estimator which means
\begin{equation*}
 0 = \delta_0^T (Y-\delta_0 \hat{\theta}-A_{ko}\gamma_{ko})
 \end{equation*}
this is actually \eqref{eq:iss2}. So plugging $\hat{\theta}$ in \eqref{eq:lasso}, we get
 \begin{eqnarray*}
 &&\|Y-\delta_0 \theta-A_{ko}\gamma_{ko}\|_2^2 \\
 &=& \|(I - \delta_0(\delta_0^T\delta_0)^{\dag}\delta_0^T)^T(Y-A_{ko}\gamma_{ko})\|_2^2\\
&= & \|U_2U_2^T(Y-A_{ko}\gamma_{ko})\|_2^2\\
 &= &  \|U_2^TY-U_2^TA_{ko}\gamma_{ko})\|_2^2.
 \end{eqnarray*}
This is in fact the $l_2$ loss for the second procedure in Remark 1. 

Finally identical paths lead to the same knockoff statistics which ends the proof.

%
%
%
%
%
%

\end{document}